\documentclass[10pt,twocolumn,letterpaper]{article}

\usepackage{cvpr}              %

\newcommand{\csquare}[1]{\raisebox{0.25 em}{{\fcolorbox{#1}{#1}{\null}}}}

\definecolor{blue_fig}{rgb}{0.165,0.373,0.659}
\definecolor{l_blue_fig}{rgb}{0.761,0.875,0.941}
\definecolor{orange_fig}{rgb}{0.941,0.788,0.529}
\definecolor{grey_fig}{rgb}{0.722,0.722,0.722}

\usepackage{tabularx}
\usepackage[table]{xcolor}
\AtBeginDocument{%
}

\renewcommand{\paragraph}[1]{%
  \vspace{4pt}%
  \par\noindent\textbf{#1}%
}

\definecolor{cvprblue}{rgb}{0.21,0.49,0.74}
\usepackage[pagebackref,breaklinks,colorlinks,allcolors=cvprblue]{hyperref}

\def\paperID{ 24} %
\def\confName{CVPR}
\def\confYear{2026}

\title{Zero-Shot Multi-Animal Tracking in the Wild}

\author{Jan F. Meier \quad Timo Lüddecke\\
Institute of Computer Science and Campus Institute Data Science\\
University of Göttingen, Germany\\
{\tt\small \{jan.meier, timo.lueddecke\}@cs.uni-goettingen.de}
}

\begin{document}
\maketitle
\begin{abstract}
Multi-animal tracking is crucial for understanding animal ecology and behavior, yet remains challenging due to variations in habitat, motion patterns, and species appearance. Traditional approaches typically require extensive fine-tuning and heuristic design for each new scenario. In this work, we explore vision foundation models for zero-shot multi-animal tracking. Building on SAM2MOT, we combine Grounding DINO with the Segment Anything Model~2 (SAM\,2) and introduce three targeted modifications to adapt the framework to animal appearance and behavior without any retraining or hyperparameter tuning between datasets. We also evaluate the recent SAM\,3 model, but identify practical limitations that restrict its applicability to multi-animal tracking in the wild. Our method achieves state-of-the-art results across Chimp\-Act, Bird Flock Tracking, AnimalTrack, and a subset of GMOT-40, demonstrating robust generalization across diverse species and environments. The code is available at \url{https://github.com/ecker-lab/SAM2-Animal-Tracking}.
\end{abstract}
    
\section{Introduction}
\label{sec:intro}
Human activity is driving an accelerating decline in animal biodiversity \cite{ceballos_accelerated_2015}, threatening ecological stability and making conservation an increasingly urgent priority \cite{ceballos_vertebrates_2020}. Evaluating the effectiveness of conservation policies requires reliable tools for monitoring animal populations at scale \cite{brookes_panaf20k_2024}. Modern recording technologies such as digital camera traps \cite{tuia_perspectives_2022} enable the collection of vast amounts of video data, but the sheer volume makes fully manual analysis infeasible \cite{farley_situating_2018}. Automated computer vision methods offer a practical path forward, and multi-animal tracking in particular provides rich information on animal presence, distribution, and behavior \cite{brookes_panaf20k_2024, zhang_animaltrack_2023, liu_deep_2024}.

Despite steady progress, most multi-animal tracking approaches remain highly domain-specific, requiring retraining for each new species, environment, or camera setup, a process that demands substantial data collection and annotation effort. This limits their applicability in practice, where labeled data is often scarce or entirely unavailable. Zero-shot models address this bottleneck directly. By leveraging vision foundation models trained on large, heterogeneous datasets, they can generalize to new domains without any retraining. This opens up valuable use cases, including generating pseudo-labels for semi-supervised training and enabling tracking in low-data scenarios.

In this work, we adapt SAM2MOT~\cite{jiang_sam2mot_2026}, a strong zero-shot tracker originally designed for human tracking, for the multi-animal setting. While SAM2MOT achieves competitive performance, it relies on manually selected detection thresholds per sequence and human-centric heuristics that do not transfer well to animal appearance and behavior. We address these limitations through three key modifications: adaptive detection thresholds that remove the need for manual tuning, mask-based track initialization that reduces spurious track creation, and density-aware reconstruction that improves robustness in crowded scenes. We additionally evaluate SAM\,3 for zero-shot animal tracking and identify practical limitations that restrict its applicability in this domain. Together, these contributions yield a tracking method that generalizes across species and environments without any retraining or dataset-specific hyperparameter tuning.

\section{Related work}
\label{sec:rel}
\vspace{-0.5em}

\begin{figure*}[tb]
  \centering
  \includegraphics[width=0.9\textwidth]{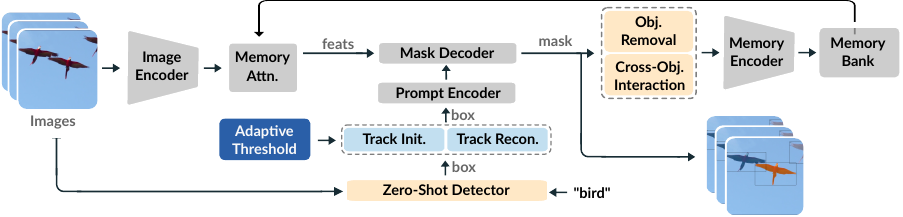}
  \caption{\textbf{Overview of our model architecture.} Gray \csquare{grey_fig} components originate from SAM 2, orange \csquare{orange_fig} components from SAM2MOT. We modify the light blue \csquare{l_blue_fig} modules from SAM2MOT and introduce the dark blue \csquare{blue_fig} ones to adapt the model for robust multi-animal tracking.}
  \label{fig:model_overview}
\vspace{-1em}
\end{figure*}

\paragraph{Multi-object tracking (MOT).}
Multi-object tracking aims to detect multiple objects in video sequences and associate their detections across frames to form consistent trajectories over time \cite{luo_multiple_2021}. Existing MOT methods can broadly be categorized into Tracking-by-Detection (TbD) and Tracking-by-Propagation (TbP) approaches \cite{gao_multiple_2025}. TbD methods first detect all objects in each frame and then associate detections through heuristic or learned strategies. Heuristic-based methods rely on motion and appearance cues for association \cite{bewley_simple_2016, wojke_simple_2017, zhang_bytetrack_2022}, while learned association models \cite{somers_cameltrack_2025} and graph-based approaches \cite{cetintas_unifying_2023, missaoui_noougat_2025} use neural networks to match detections across frames. TbP methods, inspired by the DETR architecture \cite{carion_end--end_2020}, jointly perform detection and tracking end-to-end by auto-regressively propagating object queries through time \cite{meinhardt_trackformer_2022, zeng_motr_2022}. 

Most multi-animal tracking methods fall into the heuristic TbD category \cite{pineda_deep_2022, koger_quantifying_2023, vogg_primat_2026, ma_alphachimp_2024}, with recent work targeting long-term~\cite{ngo_bibinbe_hmm-based_2026} and swarm tracking~\cite{bideau_watching_2026}. SA-FARI~\cite{wasmuht_sa_2025} finetunes and evaluates SAM\,3 for multi-animal tracking. However, SAM 3 was trained on their dataset and does not operate in a zero-shot setting, in contrast to our approach.

\paragraph{Segment Anything Models (SAM).}
SAM~\cite{kirillov_segment_2023} introduced large-scale interactive image segmentation, demonstrating strong zero-shot generalization through training on a massive curated dataset. SAM 2~\cite{ravi_sam_2025} extended this to video by maintaining a memory bank of recent masks and the initial prompt, enabling consistent object tracking and segmentation across frames given a point, bounding box, or mask prompt. Its strong zero-shot capabilities have made it a popular backbone for visual object tracking \cite{yang_samurai_2024, videnovic_distractor-aware_2025, ding_sam2long_2025}, though its application to multi-object tracking remains limited~\cite{jiang_sam2mot_2026}. The recent SAM 3~\cite{carion_sam_2026} further unifies detection, segmentation, and tracking into a single model driven by text prompts, eliminating the need for a separate detector. While SAM\,3 is able to perform multi-object tracking without modifications, we show that it has some issues which harm its applicability for multi-animal tracking (\autoref{sec:results}).

\section{Methods}
Our approach builds upon SAM2MOT \cite{jiang_sam2mot_2026}, which combines a zero-shot object detector with SAM\,2 using bounding box prompts and heuristic rules for track management. We extend this framework by reducing its reliance on hand-tuned sequence-wise detection thresholds and adapting it to the multi-animal setting (\autoref{fig:model_overview}). A zero-shot tracker should ideally generalize to unseen data without any hyperparameter tuning. To this end, we introduce three targeted but intentionally simple modifications to SAM2MOT. \textbf{Adaptive detection thresholds} automatically adjust detection confidence based on scene statistics, removing the need for manual threshold selection per sequence. \textbf{Mask-based track initialization} uses SAM\,2 segmentation masks rather than bounding boxes alone to decide whether a detection corresponds to a new object, reducing false track initializations. \textbf{Density-aware reconstruction} limits re-prompting of existing tracks to spatially unambiguous detections, improving robustness in crowded scenes. We additionally apply non-maximum suppression (NMS) to the track masks to suppress duplicate tracks. These modifications are motivated by challenges common in animal tracking: highly variable detection score distributions across species and habitats, frequent occlusion, and dense clustering from herding and flocking behavior.

\begin{figure}[htb]
  \centering
  \includegraphics[width=\linewidth]{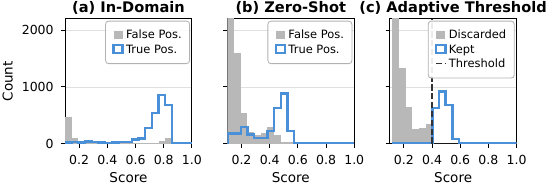}
  \caption{\textbf{Detection score distributions and adaptive thresholding.} (a) In-domain detector scores, showing a broad range of effective thresholds. (b) Zero-shot detector scores, with higher sensitivity to threshold selection. (c) Detection assignment using K-Means–based adaptive thresholding, which automatically separates true positives from false positives without manual tuning.}
  \label{fig:dist_fig}
\vspace{-0.5em}
\end{figure}

\paragraph{Adaptive detection thresholds.}
Detection score distributions differ markedly between in-domain (trained on target dataset) and zero-shot detectors and vary significantly across sequences (\autoref{fig:dist_fig}, \autoref{sec:appendix_dist}). While in-domain detectors typically exhibit a broad range of effective thresholds, zero-shot detectors are far more sensitive to threshold choice, making a fixed global threshold insufficient. The standard remedy of sweeping thresholds and selecting the best-performing one is only feasible when labeled data is available, and therefore incompatible with a truly zero-shot setting.

We address this by proposing an adaptive thresholding method based on K-Means clustering of the per-sequence detection scores. The scores are partitioned into two clusters, with the lower-scoring cluster treated as false positives and the higher-scoring cluster as true positives. The threshold is then set to the decision boundary between the two clusters. This connects to classical adaptive thresholding \cite{otsu_threshold_1979}, applied here to per-sequence zero-shot detection scores. Since our model relies on detections primarily for track initialization and lacks semantic class understanding, false positive detections are particularly harmful. We therefore add a small static offset $\delta$ to the adaptive threshold to favor high precision; $\delta$ is fixed across all datasets and sequences.

\paragraph{Mask-based track initialization.}
To decide whether a detection corresponds to an existing track or a new object, SAM2MOT computes the fraction of pixels within the detection bounding box that are not assigned to any active track. This heuristic can fail when bounding boxes overlap. If an animal is partially occluded by another one, most pixels in its bounding box may already be assigned to the foreground track, causing the system to miss the new object. Box-mask overlap alone cannot distinguish this from a redundant detection of the foreground animal. To address this, we prompt SAM\,2 with the detection bounding box to obtain a segmentation mask $M_{\text{det}}$. For each active track with mask $M_i$, we compute the \emph{mask overlap ratio}, defined as
\begin{equation}
\frac{\lvert M_{\text{det}} \cap M_i \rvert}{\lvert M_{\text{det}} \rvert}.
\end{equation}
A new track is initialized only if the maximum mask overlap ratio across all active tracks is below a threshold $\tau_{\text{mask}}$, indicating that the detection is unlikely to correspond to any existing track. This mask-level criterion is more robust to occlusion than the pixel-assignment heuristic it replaces.

\paragraph{Density-aware reconstruction.}
Track mask quality degrades over time, and SAM2MOT counteracts this by re-initializing tracks using detection bounding boxes. However, detection quality deteriorates in crowded scenes \cite{zheng_progressive_2022}, and re-initializing with a low-quality box can introduce noise rather than improve the track. We therefore restrict re-initialization to spatially unambiguous detections. To identify such cases, we compare the IoU between the detection box and the bounding box of each track mask. A detection is considered unambiguous only if the gap between the highest and second-highest IoU exceeds a threshold $\tau_{\text{IoU}}$, ensuring a clear one-to-one correspondence between the detection and a single existing track.

\section{Experiments}
\label{sec:exp}
\vspace{-0.5em}

\paragraph{Datasets and metrics.}
To demonstrate broad applicability, we evaluate our method on four animal tracking datasets covering different species and scenarios. ChimpAct~\cite{ma_chimpact_2023} contains videos of chimpanzees from the Leipzig Zoo, featuring relatively few animals per frame. Bird Flock Tracking (BFT)~\cite{zheng_nettrack_2024} includes sequences of 22 bird species in diverse environments, focusing on highly dynamic objects. AnimalTrack~\cite{zhang_animaltrack_2023} comprises 10 common animal categories with many individual tracks per sequence. GMOT-40-Animal is a subset of GMOT-40~\cite{bai_gmot-40_2021} containing 16 sequences with animals including birds, fish, insects, and livestock. Extended dataset details are provided in \autoref{sec:appendix_dataset}.

We evaluate performance using three standard multi-object tracking metrics: Higher Order Tracking Accuracy (HOTA)~\cite{luiten_hota_2021}, Detection Accuracy (DetA), and Association Accuracy (AssA), focusing primarily on HOTA as a unified metric. DetA and AssA are reported separately to assess detection and association performance.

\paragraph{Implementation details.}
For detection, we use the Grounding DINO Swin-L~\cite{liu_grounding_2024} checkpoint provided by MMDetection~\cite{chen_mmdetection_2019}, and for propagation, SAM\,2.1 Hiera-L from the official repository~\cite{ravi_sam_2025}. The same model checkpoints and hyperparameters are used across all datasets (\autoref{sec:appendix_hyperparam}). Evaluation is conducted using TrackEval~\cite{jonathon_luiten_trackeval_2020}. All experiments are run on a single NVIDIA A100/H100 GPU.

\section{Results}
\label{sec:results}
We compare our method against state-of-the-art approaches across four benchmark datasets (\autoref{tab:results}). 
It performs comparably to SAM 3 on ChimpAct and BFT and outperforms it on GMOT-40-Animal, while avoiding its practical limitations in memory, multi-class support, and prompt sensitivity. Notably, our method considerably outperforms other zero-shot methods using the same detections as well as models trained on the respective dataset. It demonstrates its robustness and generalization ability in diverse tracking scenarios without dataset-specific training. Visualizations and failure cases are in \autoref{sec:failure_cases} and \ref{sec:vis}.

\begin{table*}[htb]
  \caption{\textbf{State-of-the-art comparison on multiple multi-animal tracking datasets.} Results are reported on the test set. The performance of in-domain methods was taken from the respective dataset or method paper. Our method outperforms both trained and zero-shot approaches using the same detections. GMOT-40-Animal provides only a test split, precluding training on the target dataset entirely. ``out-of-memory'' indicates the model ran out of memory on an NVIDIA H100 during inference. Extended metrics are reported in \autoref{sec:appendix_metrics}.}
  \vspace{-0.5em}
  \label{tab:results}
  \centering
  \footnotesize
  \setlength{\tabcolsep}{5pt}
  \begin{tabular*}{\linewidth}{@{\extracolsep{\fill}}llcccccccccccc}
    \toprule
    & & \multicolumn{3}{c}{ChimpAct} & \multicolumn{3}{c}{BFT}
      & \multicolumn{3}{c}{AnimalTrack} & \multicolumn{3}{c}{GMOT-40-Animal} \\
    \cmidrule(lr){3-5} \cmidrule(lr){6-8} \cmidrule(lr){9-11} \cmidrule(lr){12-14}
    Model & Detector
      & HOTA & DetA & AssA
      & HOTA & DetA & AssA
      & HOTA & DetA & AssA
      & HOTA & DetA & AssA \\
    \midrule
    \multicolumn{14}{l}{\textit{Trained on target dataset}} \\
    \midrule
    ByteTrack~\cite{zhang_bytetrack_2022}
      & YOLO-X~\cite{ge_yolox_2021}
      & 49.2 & $-$ & $-$ & 52.5 & 51.6 & 53.7 & 40.1 & $-$ & $-$ & $-$ & $-$ & $-$ \\
    AlphaChimp~\cite{ma_alphachimp_2024}
      & Custom DETR
      & 56.3 & $-$ & $-$ & $-$ & $-$ & $-$ & $-$ & $-$ & $-$ & $-$ & $-$ & $-$ \\
    MOTIP~\cite{gao_multiple_2025}
      & Def. DETR~\cite{zhu_deformable_2021}
      & $-$ & $-$ & $-$ & 70.5 & 69.6 & 71.8 & $-$ & $-$ & $-$ & $-$ & $-$ & $-$ \\
    \midrule
    \multicolumn{14}{l}{\textit{Zero-shot}} \\
    \midrule
    ByteTrack~\cite{zhang_bytetrack_2022}
      & Grounding DINO L
      & 50.6 & 45.8 & 57.1 & 59.0 & 60.4 & 57.9 & 48.3 & 43.6 & 54.1 & 48.4 & 41.7 & 56.6 \\
    NetTrack~\cite{zheng_nettrack_2024}
      & Grounding DINO L
      & 49.2 & 44.2 & 55.4 & 68.4 & 70.4 & 66.7 & 48.2 & 45.1 & 52.1 & 43.2 & 46.6 & 40.4 \\
    SAM\,3~\cite{carion_sam_2026}
      & SAM\,3
      & \textbf{58.9} & 48.5 & \textbf{72.5}
      & \textbf{75.4} & 71.5 & \textbf{79.7}
      & \multicolumn{3}{c}{\textit{out-of-memory}}
      & 60.4 & 52.6 & \textbf{70.6} \\
    \textbf{Ours}
      & Grounding DINO L
      & 58.6 & \textbf{49.8} & 70.1
      & 74.8 & \textbf{72.2} & 77.7
      & \textbf{58.0} & \textbf{52.7} & \textbf{65.2}
      & \textbf{62.4} & \textbf{57.2} & 69.2 \\
    \bottomrule
  \end{tabular*}
\vspace{-0.5em}
\end{table*}

\paragraph{Effect of added components.} 
We ablate on ChimpAct and BFT, as these are the only datasets with dedicated validation splits. We demonstrate that each proposed component consistently improves tracking performance (\autoref{tab:main_ablation}). Adaptive detection thresholds increase both detection and association accuracy, mask-based track initialization further strengthens associations, and density-aware reconstruction improves robustness in crowded scenes. Mask NMS reduces false positives and offers gains in HOTA on ChimpAct. On BFT, it reduces false positives (3014 to 2794), but also removes valid tracks in dense, fast-moving flocks that would have been correctly reassociated, lowering AssA and resulting in a net HOTA decrease. We retain it as a unified design choice, as it benefits the remaining datasets and avoids per-dataset tuning.
\begin{table}[htb]
  \caption{\textbf{Our modifications improve performance.} Ablation of each component on the ChimpAct and BFT validation splits.}
  \vspace{-0.5em}
  \label{tab:main_ablation}
  \centering
  \footnotesize
  \setlength{\tabcolsep}{3.5pt}
  \begin{tabular*}{\linewidth}{@{\extracolsep{\fill}}lcccccc}
    \toprule
    & \multicolumn{3}{c}{ChimpAct} & \multicolumn{3}{c}{BFT}\\
    \cmidrule(lr){2-4} \cmidrule(lr){5-7}
    Components & HOTA & DetA & AssA & HOTA & DetA & AssA\\
    \midrule
    Baseline & 54.6 & 44.5 & 67.7 & 71.6 & 68.1 & 75.4 \\
    + Adapt. det. th.   & 56.7 & 46.4 & 69.9 & 72.2 & 69.3 & 75.5 \\
    + Mask track init. & 56.9 & 46.8 & 70.0 & 72.4 & 69.4 & 75.7 \\
    + Density-aware rec.  & 57.0 & 46.7 & \textbf{70.4} & \textbf{73.2} & 70.2 & \textbf{76.5} \\
    + Mask NMS & \textbf{57.3} & \textbf{47.3} & 70.1 & 72.5 & \textbf{70.7} & 74.6 \\
    \bottomrule
  \end{tabular*}
\vspace{-0.5em}
\end{table}
Our proposed adaptive detection threshold method achieves significant performance gains compared to fixed thresholds (\autoref{tab:adaptive_th}). We attribute this improvement to variations in detection score distributions across sequences within the same dataset. Since each sequence may have a different optimal detection threshold, a single global threshold cannot adapt effectively. Our method addresses this by generating a threshold for each sequence, improving tracking performance.
\begin{table}[htb]
  \caption{\textbf{Per-sequence adaptive detection thresholds improve performance over global thresholds.} Results are reported on the val splits; the baseline is our model without the proposed changes.}
  \vspace{-0.5em}
  \label{tab:adaptive_th}
  \centering
  \footnotesize
  \setlength{\tabcolsep}{3.5pt}
  \begin{tabular*}{\linewidth}{@{\extracolsep{\fill}}ccccccc}
    \toprule
     & \multicolumn{3}{c}{ChimpAct} & \multicolumn{3}{c}{BFT}\\
    \cmidrule(lr){2-4} \cmidrule(lr){5-7}
    Det. Threshold  & HOTA & DetA & AssA & HOTA & DetA & AssA\\
    \midrule
    0.3 & 46.5 & 36.8 & 59.8 & 57.7 & 48.9 & 68.4 \\
    0.4 & 54.6 & 44.5 & 67.7 & 70.9 & 67.4 & 74.8 \\
    0.5 & 54.0 & 43.9 & 67.0 & 71.6 & 68.1 & 75.4 \\
    Adaptive & \textbf{56.7} & \textbf{46.4} & \textbf{69.9} & \textbf{72.2} & \textbf{69.3} & \textbf{75.5} \\
    \bottomrule
  \end{tabular*}
\vspace{-0.5em}
\end{table}

\begin{figure}
    \centering
    \includegraphics[width=\linewidth]{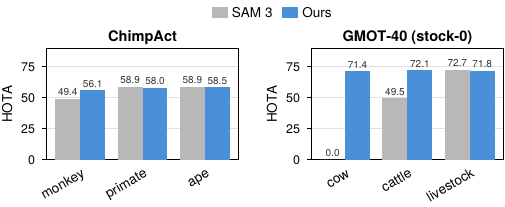}
    \caption{\textbf{SAM 3 performance is sensitive to the text prompt.} Different text prompts significantly affect SAM 3's performance, whereas our method is more robust.}
    \label{fig:text_sens}
\vspace{-1em}
\end{figure}

\paragraph{Comparison to SAM\,3.}
At first glance, SAM\,3 appears to be a straightforward upgrade, unifying detection and tracking in a single model pass instead of the two separate forward passes our pipeline requires. However, we identify three practical limitations that restrict its applicability to multi-animal tracking. First, SAM\,3 has substantial memory requirements. Even with \texttt{offload\_video\_to\_cpu} and \texttt{offload\_state\_to\_cpu} enabled, it runs out of memory on an NVIDIA H100 with 96 GB of VRAM for the AnimalTrack sequences \texttt{duck\_3}, \texttt{goose\_3}, \texttt{horse\_3}, and \texttt{rabbit\_2}. Beyond memory, SAM\,3 must be run $n$ times for $n$ text prompts, as it does not support multi-class tracking in a single pass. This not only increases runtime but also prevents heuristics from resolving duplicate detections across passes, whereas our method handles multiple prompts in a single pass by design. Finally, the SAM\,3 text encoder is sensitive to exact prompt wording~\cite{archit_revisiting_2026}, as illustrated for ChimpAct and GMOT-40-Animal (\autoref{fig:text_sens}). This means that identifying a working prompt requires manual inspection, undermining the zero-shot premise.

\section{Conclusion}
\label{sec:conc}
We analyze the performance of various zero-shot multi-animal trackers on four benchmarks and present an adaptation of SAM2MOT for zero-shot multi-animal tracking, introducing heuristics tailored to animal behavior and scene complexity. Our method leverages foundation models to deliver robust tracking performance across diverse species and environments without retraining or hyperparameter tuning between datasets. Evaluations show consistent improvements in detection and association accuracy, demonstrating the promise of zero-shot tracking for scalable wildlife monitoring and behavioral analysis. Limitations are long per-frame runtime and reduced scalability when the number of simultaneous tracks is large, as both runtime and memory requirements scale linearly with the number of active tracks (\autoref{sec:appendix_runtime}).

\section*{Acknowledgments}
The project was funded by the Deutsche Forschungsgemeinschaft (DFG, German Research Foundation) – Project-ID 454648639 – SFB 1528. The authors gratefully acknowledge the computing time granted by the Resource Allocation Board and provided on the supercomputer Emmy/Grete at NHR-Nord@G\"ottingen as part of the NHR infrastructure. The calculations for this research were conducted with computing resources under the project nib00021.

{
    \small
    \bibliographystyle{ieeenat_fullname}
    \bibliography{main.bib}

\begin{thebibliography}{44}
\providecommand{\natexlab}[1]{#1}
\providecommand{\url}[1]{\texttt{#1}}
\expandafter\ifx\csname urlstyle\endcsname\relax
  \providecommand{\doi}[1]{doi: #1}\else
  \providecommand{\doi}{doi: \begingroup \urlstyle{rm}\Url}\fi

\bibitem[Archit and Pape(2026)]{archit_revisiting_2026}
Anwai Archit and Constantin Pape.
\newblock Revisiting foundation models for cell instance segmentation.
\newblock In \emph{Medical Imaging with Deep Learning}, 2026.

\bibitem[Bai et~al.(2021)Bai, Cheng, Chu, Liu, Zhang, and
  Ling]{bai_gmot-40_2021}
Hexin Bai, Wensheng Cheng, Peng Chu, Juehuan Liu, Kai Zhang, and Haibin Ling.
\newblock Gmot-40: {A} benchmark for generic multiple object tracking.
\newblock In \emph{CVPR}, pages 6719--6728, 2021.

\bibitem[Bewley et~al.(2016)Bewley, Ge, Ott, Ramos, and
  Upcroft]{bewley_simple_2016}
Alex Bewley, Zongyuan Ge, Lionel Ott, Fabio Ramos, and Ben Upcroft.
\newblock Simple online and realtime tracking.
\newblock In \emph{ICIP}, pages 3464--3468. IEEE, 2016.

\bibitem[Bideau et~al.(2026)Bideau, Pham, Dhellemmes, Hansen, and
  Krause]{bideau_watching_2026}
Pia Bideau, Duc Pham, Félicie Dhellemmes, Matthew Hansen, and Jens Krause.
\newblock Watching {Swarm} {Dynamics} from {Above}: {A} {Framework} for
  {Advanced} {Object} {Tracking} in {Drone} {Videos}.
\newblock \emph{IJCV}, 134\penalty0 (4):\penalty0 146, 2026.

\bibitem[Brookes et~al.(2024)Brookes, Mirmehdi, Stephens, Angedakin, Corogenes,
  Dowd, Dieguez, Hicks, Jones, Lee, Leinert, Lapuente, McCarthy, Meier, Murai,
  Normand, Vergnes, Wessling, Wittig, Langergraber, Maldonado, Yang,
  Zuberbühler, Boesch, Arandjelovic, Kühl, and
  Burghardt]{brookes_panaf20k_2024}
Otto Brookes, Majid Mirmehdi, Colleen Stephens, Samuel Angedakin, Katherine
  Corogenes, Dervla Dowd, Paula Dieguez, Thurston~C. Hicks, Sorrel Jones, Kevin
  Lee, Vera Leinert, Juan Lapuente, Maureen~S. McCarthy, Amelia Meier, Mizuki
  Murai, Emmanuelle Normand, Virginie Vergnes, Erin~G. Wessling, Roman~M.
  Wittig, Kevin Langergraber, Nuria Maldonado, Xinyu Yang, Klaus Zuberbühler,
  Christophe Boesch, Mimi Arandjelovic, Hjalmar Kühl, and Tilo Burghardt.
\newblock {PanAf20K}: {A} {Large} {Video} {Dataset} for {Wild} {Ape}
  {Detection} and {Behaviour} {Recognition}.
\newblock \emph{IJCV}, 132\penalty0 (8):\penalty0 3086--3102, 2024.

\bibitem[Carion et~al.(2020)Carion, Massa, Synnaeve, Usunier, Kirillov, and
  Zagoruyko]{carion_end--end_2020}
Nicolas Carion, Francisco Massa, Gabriel Synnaeve, Nicolas Usunier, Alexander
  Kirillov, and Sergey Zagoruyko.
\newblock End-to-end object detection with transformers.
\newblock In \emph{ECCV}, pages 213--229. Springer, 2020.

\bibitem[Carion et~al.(2026)Carion, Gustafson, Hu, Debnath, Hu, Coll-Vinent,
  Ryali, Alwala, Khedr, Huang, Lei, Ma, Guo, Kalla, Marks, Greer, Wang, Sun,
  R{\"a}dle, Afouras, Mavroudi, Xu, Wu, Zhou, Momeni, Hazra, Ding, Vaze,
  Porcher, Li, Li, Kamath, Cheng, Dollar, Ravi, Saenko, Zhang, and
  Feichtenhofer]{carion_sam_2026}
Nicolas Carion, Laura Gustafson, Yuan-Ting Hu, Shoubhik Debnath, Ronghang Hu,
  Didac~Suris Coll-Vinent, Chaitanya Ryali, Kalyan~Vasudev Alwala, Haitham
  Khedr, Andrew Huang, Jie Lei, Tengyu Ma, Baishan Guo, Arpit Kalla, Markus
  Marks, Joseph Greer, Meng Wang, Peize Sun, Roman R{\"a}dle, Triantafyllos
  Afouras, Effrosyni Mavroudi, Katherine Xu, Tsung-Han Wu, Yu Zhou, Liliane
  Momeni, Rishi Hazra, Shuangrui Ding, Sagar Vaze, Francois Porcher, Feng Li,
  Siyuan Li, Aishwarya Kamath, Ho~Kei Cheng, Piotr Dollar, Nikhila Ravi, Kate
  Saenko, Pengchuan Zhang, and Christoph Feichtenhofer.
\newblock {SAM} 3: Segment anything with concepts.
\newblock In \emph{ICLR}, 2026.

\bibitem[Ceballos et~al.(2015)Ceballos, Ehrlich, Barnosky, García, Pringle,
  and Palmer]{ceballos_accelerated_2015}
Gerardo Ceballos, Paul~R. Ehrlich, Anthony~D. Barnosky, Andrés García,
  Robert~M. Pringle, and Todd~M. Palmer.
\newblock Accelerated modern human–induced species losses: {Entering} the
  sixth mass extinction.
\newblock \emph{Science Advances}, 2015.
\newblock Publisher: American Association for the Advancement of Science.

\bibitem[Ceballos et~al.(2020)Ceballos, Ehrlich, and
  Raven]{ceballos_vertebrates_2020}
Gerardo Ceballos, Paul~R. Ehrlich, and Peter~H. Raven.
\newblock Vertebrates on the brink as indicators of biological annihilation and
  the sixth mass extinction.
\newblock \emph{Proceedings of the National Academy of Sciences}, 117\penalty0
  (24):\penalty0 13596--13602, 2020.

\bibitem[Cetintas et~al.(2023)Cetintas, Brasó, and
  Leal-Taixé]{cetintas_unifying_2023}
Orcun Cetintas, Guillem Brasó, and Laura Leal-Taixé.
\newblock Unifying short and long-term tracking with graph hierarchies.
\newblock In \emph{CVPR}, pages 22877--22887, 2023.

\bibitem[Chen et~al.(2019)Chen, Wang, Pang, Cao, Xiong, Li, Sun, Feng, Liu, Xu,
  Zhang, Cheng, Zhu, Cheng, Zhao, Li, Lu, Zhu, Wu, Dai, Wang, Shi, Ouyang, Loy,
  and Lin]{chen_mmdetection_2019}
Kai Chen, Jiaqi Wang, Jiangmiao Pang, Yuhang Cao, Yu Xiong, Xiaoxiao Li,
  Shuyang Sun, Wansen Feng, Ziwei Liu, Jiarui Xu, Zheng Zhang, Dazhi Cheng,
  Chenchen Zhu, Tianheng Cheng, Qijie Zhao, Buyu Li, Xin Lu, Rui Zhu, Yue Wu,
  Jifeng Dai, Jingdong Wang, Jianping Shi, Wanli Ouyang, Chen~Change Loy, and
  Dahua Lin.
\newblock {MMDetection}: {Open} {MMLab} {Detection} {Toolbox} and {Benchmark}.
\newblock \emph{arXiv preprint arXiv:1906.07155}, 2019.

\bibitem[Ding et~al.(2025)Ding, Qian, Dong, Zhang, Zang, Cao, Guo, Lin, and
  Wang]{ding_sam2long_2025}
Shuangrui Ding, Rui Qian, Xiaoyi Dong, Pan Zhang, Yuhang Zang, Yuhang Cao,
  Yuwei Guo, Dahua Lin, and Jiaqi Wang.
\newblock Sam2long: Enhancing sam 2 for long video segmentation with a
  training-free memory tree.
\newblock In \emph{ICCV}, pages 13614--13624, 2025.

\bibitem[Farley et~al.(2018)Farley, Dawson, Goring, and
  Williams]{farley_situating_2018}
Scott~S Farley, Andria Dawson, Simon~J Goring, and John~W Williams.
\newblock Situating {Ecology} as a {Big}-{Data} {Science}: {Current}
  {Advances}, {Challenges}, and {Solutions}.
\newblock \emph{BioScience}, 68\penalty0 (8):\penalty0 563--576, 2018.

\bibitem[Gao et~al.(2025)Gao, Qi, and Wang]{gao_multiple_2025}
Ruopeng Gao, Ji Qi, and Limin Wang.
\newblock Multiple object tracking as id prediction.
\newblock In \emph{CVPR}, pages 27883--27893, 2025.

\bibitem[Ge et~al.(2021)Ge, Liu, Wang, Li, and Sun]{ge_yolox_2021}
Zheng Ge, Songtao Liu, Feng Wang, Zeming Li, and Jian Sun.
\newblock {YOLOX}: {Exceeding} {YOLO} {Series} in 2021.
\newblock \emph{arXiv preprint arXiv:2107.08430}, 2021.

\bibitem[Jiang et~al.(2026)Jiang, Wang, Zhao, Li, and
  Jiang]{jiang_sam2mot_2026}
Junjie Jiang, Zelin Wang, Manqi Zhao, Yin Li, and DongSheng Jiang.
\newblock Sam2mot: A novel paradigm of multi-object tracking by segmentation.
\newblock In \emph{AAAI}, pages 5388--5396, 2026.

\bibitem[Kirillov et~al.(2023)Kirillov, Mintun, Ravi, Mao, Rolland, Gustafson,
  Xiao, Whitehead, Berg, Lo, et~al.]{kirillov_segment_2023}
Alexander Kirillov, Eric Mintun, Nikhila Ravi, Hanzi Mao, Chloe Rolland, Laura
  Gustafson, Tete Xiao, Spencer Whitehead, Alexander~C Berg, Wan-Yen Lo, et~al.
\newblock Segment anything.
\newblock In \emph{ICCV}, pages 4015--4026, 2023.

\bibitem[Koger et~al.(2023)Koger, Deshpande, Kerby, Graving, Costelloe, and
  Couzin]{koger_quantifying_2023}
Benjamin Koger, Adwait Deshpande, Jeffrey~T. Kerby, Jacob~M. Graving, Blair~R.
  Costelloe, and Iain~D. Couzin.
\newblock Quantifying the movement, behaviour and environmental context of
  group-living animals using drones and computer vision.
\newblock \emph{Journal of Animal Ecology}, 92\penalty0 (7):\penalty0
  1357--1371, 2023.

\bibitem[Liu et~al.(2024{\natexlab{a}})Liu, Zeng, Ren, Li, Zhang, Yang, Jiang,
  Li, Yang, Su, Zhu, and Zhang]{liu_grounding_2024}
Shilong Liu, Zhaoyang Zeng, Tianhe Ren, Feng Li, Hao Zhang, Jie Yang, Qing
  Jiang, Chunyuan Li, Jianwei Yang, Hang Su, Jun Zhu, and Lei Zhang.
\newblock Grounding {DINO}: {Marrying} {DINO} with {Grounded} {Pre}-{Training}
  for {Open}-{Set} {Object} {Detection}, 2024{\natexlab{a}}.
\newblock arXiv:2303.05499 [cs].

\bibitem[Liu et~al.(2024{\natexlab{b}})Liu, Li, Liu, Li, and
  Yue]{liu_deep_2024}
Yeqiang Liu, Weiran Li, Xue Liu, Zhenbo Li, and Jun Yue.
\newblock Deep learning in multiple animal tracking: {A} survey.
\newblock \emph{Computers and Electronics in Agriculture}, 224:\penalty0
  109161, 2024{\natexlab{b}}.

\bibitem[Luiten and Hoffhues(2020)]{jonathon_luiten_trackeval_2020}
Jonathon Luiten and Arne Hoffhues.
\newblock {TrackEval}, 2020.

\bibitem[Luiten et~al.(2021)Luiten, Osep, Dendorfer, Torr, Geiger, Leal-Taixé,
  and Leibe]{luiten_hota_2021}
Jonathon Luiten, Aljosa Osep, Patrick Dendorfer, Philip Torr, Andreas Geiger,
  Laura Leal-Taixé, and Bastian Leibe.
\newblock Hota: {A} higher order metric for evaluating multi-object tracking.
\newblock \emph{IJCV}, 129\penalty0 (2):\penalty0 548--578, 2021.
\newblock Publisher: Springer.

\bibitem[Luo et~al.(2021)Luo, Xing, Milan, Zhang, Liu, and
  Kim]{luo_multiple_2021}
Wenhan Luo, Junliang Xing, Anton Milan, Xiaoqin Zhang, Wei Liu, and Tae-Kyun
  Kim.
\newblock Multiple object tracking: {A} literature review.
\newblock \emph{Artificial Intelligence}, 293:\penalty0 103448, 2021.

\bibitem[Ma et~al.(2023)Ma, Kaufhold, Su, Zhu, Terwilliger, Meza, Zhu, Rossano,
  and Wang]{ma_chimpact_2023}
Xiaoxuan Ma, Stephan Kaufhold, Jiajun Su, Wentao Zhu, Jack Terwilliger, Andres
  Meza, Yixin Zhu, Federico Rossano, and Yizhou Wang.
\newblock Chimpact: {A} longitudinal dataset for understanding chimpanzee
  behaviors.
\newblock \emph{NeurIPS}, 36:\penalty0 27501--27531, 2023.

\bibitem[Ma et~al.(2024)Ma, Lin, Xu, Kaufhold, Terwilliger, Meza, Zhu, Rossano,
  and Wang]{ma_alphachimp_2024}
Xiaoxuan Ma, Yutang Lin, Yuan Xu, Stephan~P Kaufhold, Jack Terwilliger, Andres
  Meza, Yixin Zhu, Federico Rossano, and Yizhou Wang.
\newblock {AlphaChimp}: {Tracking} and {Behavior} {Recognition} of
  {Chimpanzees}.
\newblock \emph{arXiv preprint arXiv:2410.17136}, 2024.

\bibitem[Meinhardt et~al.(2022)Meinhardt, Kirillov, Leal-Taixe, and
  Feichtenhofer]{meinhardt_trackformer_2022}
Tim Meinhardt, Alexander Kirillov, Laura Leal-Taixe, and Christoph
  Feichtenhofer.
\newblock Trackformer: {Multi}-object tracking with transformers.
\newblock In \emph{CVPR}, pages 8844--8854, 2022.

\bibitem[Missaoui et~al.(2025)Missaoui, Cetintas, Brasó, Meinhardt, and
  Leal-Taixé]{missaoui_noougat_2025}
Benjamin Missaoui, Orcun Cetintas, Guillem Brasó, Tim Meinhardt, and Laura
  Leal-Taixé.
\newblock {NOOUGAT}: {Towards} {Unified} {Online} and {Offline}
  {Multi}-{Object} {Tracking}.
\newblock \emph{arXiv preprint arXiv:2509.02111}, 2025.

\bibitem[Ngo~Bibinbe et~al.(2026)Ngo~Bibinbe, Bang, Gagnon, Ahloy-Dallaire, and
  Paquet]{ngo_bibinbe_hmm-based_2026}
Anne Marthe~Sophie Ngo~Bibinbe, Chiron Bang, Patrick Gagnon, Jamie
  Ahloy-Dallaire, and Eric~R. Paquet.
\newblock An {HMM}-{Based} {Framework} for {Identity}-{Aware} {Long}-{Term}
  {Multi}-{Object} {Tracking} {From} {Sparse} and {Uncertain} {Identification}:
  {Use} {Case} on {Long}-{Term} {Tracking} in {Livestock}.
\newblock \emph{IJCV}, 134\penalty0 (3):\penalty0 107, 2026.

\bibitem[Otsu(1979)]{otsu_threshold_1979}
Nobuyuki Otsu.
\newblock A {Threshold} {Selection} {Method} from {Gray}-{Level} {Histograms}.
\newblock \emph{IEEE Transactions on Systems, Man, and Cybernetics}, 9\penalty0
  (1):\penalty0 62--66, 1979.

\bibitem[Pineda et~al.(2022)Pineda, Kubo, Shimada, and Ikeda]{pineda_deep_2022}
Riza~Rae Pineda, Takatomi Kubo, Masaki Shimada, and Kazushi Ikeda.
\newblock Deep {MAnTra}: deep learning-based multi-animal tracking for
  {Japanese} macaques.
\newblock \emph{Artif. Life Robot.}, 28\penalty0 (1):\penalty0 127--138, 2022.

\bibitem[Ravi et~al.(2025)Ravi, Gabeur, Hu, Hu, Ryali, Ma, Khedr, R{\"a}dle,
  Rolland, Gustafson, Mintun, Pan, Alwala, Carion, Wu, Girshick, Dollar, and
  Feichtenhofer]{ravi_sam_2025}
Nikhila Ravi, Valentin Gabeur, Yuan-Ting Hu, Ronghang Hu, Chaitanya Ryali,
  Tengyu Ma, Haitham Khedr, Roman R{\"a}dle, Chloe Rolland, Laura Gustafson,
  Eric Mintun, Junting Pan, Kalyan~Vasudev Alwala, Nicolas Carion, Chao-Yuan
  Wu, Ross Girshick, Piotr Dollar, and Christoph Feichtenhofer.
\newblock {SAM} 2: Segment anything in images and videos.
\newblock In \emph{ICLR}, 2025.

\bibitem[Somers et~al.(2025)Somers, Standaert, Joos, Alahi, and
  De~Vleeschouwer]{somers_cameltrack_2025}
Vladimir Somers, Baptiste Standaert, Victor Joos, Alexandre Alahi, and
  Christophe De~Vleeschouwer.
\newblock {CAMELTrack}: {Context}-{Aware} {Multi}-cue {ExpLoitation} for
  {Online} {Multi}-{Object} {Tracking}.
\newblock \emph{arXiv preprint arXiv:2505.01257}, 2025.

\bibitem[Tuia et~al.(2022)Tuia, Kellenberger, Beery, Costelloe, Zuffi, Risse,
  Mathis, Mathis, van Langevelde, Burghardt, Kays, Klinck, Wikelski, Couzin,
  van Horn, Crofoot, Stewart, and Berger-Wolf]{tuia_perspectives_2022}
Devis Tuia, Benjamin Kellenberger, Sara Beery, Blair~R. Costelloe, Silvia
  Zuffi, Benjamin Risse, Alexander Mathis, Mackenzie~W. Mathis, Frank van
  Langevelde, Tilo Burghardt, Roland Kays, Holger Klinck, Martin Wikelski,
  Iain~D. Couzin, Grant van Horn, Margaret~C. Crofoot, Charles~V. Stewart, and
  Tanya Berger-Wolf.
\newblock Perspectives in machine learning for wildlife conservation.
\newblock \emph{Nature Communications}, 13\penalty0 (1):\penalty0 792, 2022.
\newblock Publisher: Nature Publishing Group.

\bibitem[Videnovic et~al.(2025)Videnovic, Lukezic, and
  Kristan]{videnovic_distractor-aware_2025}
Jovana Videnovic, Alan Lukezic, and Matej Kristan.
\newblock A distractor-aware memory for visual object tracking with sam2.
\newblock In \emph{CVPR}, pages 24255--24264, 2025.

\bibitem[Vogg et~al.(2026)Vogg, Nuske, Weis, Lüddecke, Karakoç, Ahmed,
  Pereira, Malaivijitnond, Meesawat, Murphy, Fischer, Wörgötter, Kappeler,
  Gail, Ostner, Schülke, Fichtel, and Ecker]{vogg_primat_2026}
Richard Vogg, Matthias Nuske, Marissa~A. Weis, Timo Lüddecke, Elif Karakoç,
  Zurna Ahmed, Sofia~M. Pereira, Suchinda Malaivijitnond, Suthirote Meesawat,
  Derek Murphy, Julia Fischer, Florentin Wörgötter, Peter~M. Kappeler,
  Alexander Gail, Julia Ostner, Oliver Schülke, Claudia Fichtel, and
  Alexander~S. Ecker.
\newblock {PriMAT}: {Robust} multi-animal tracking of primates in the wild.
\newblock \emph{PLOS ONE}, 21\penalty0 (4):\penalty0 e0347669, 2026.

\bibitem[Wasmuht et~al.(2025)Wasmuht, Brookes, Schall, Palencia, Beirne,
  Burghardt, Mirmehdi, K{\"u}hl, Arandjelovic, Pottie, et~al.]{wasmuht_sa_2025}
Dante~Francisco Wasmuht, Otto Brookes, Maximillian Schall, Pablo Palencia,
  Chris Beirne, Tilo Burghardt, Majid Mirmehdi, Hjalmar K{\"u}hl, Mimi
  Arandjelovic, Sam Pottie, et~al.
\newblock The sa-fari dataset: Segment anything in footage of animals for
  recognition and identification.
\newblock \emph{arXiv preprint arXiv:2511.15622}, 2025.

\bibitem[Wojke et~al.(2017)Wojke, Bewley, and Paulus]{wojke_simple_2017}
Nicolai Wojke, Alex Bewley, and Dietrich Paulus.
\newblock Simple online and realtime tracking with a deep association metric.
\newblock In \emph{ICIP}, pages 3645--3649. IEEE, 2017.

\bibitem[Yang et~al.(2024)Yang, Huang, Chai, Jiang, and
  Hwang]{yang_samurai_2024}
Cheng-Yen Yang, Hsiang-Wei Huang, Wenhao Chai, Zhongyu Jiang, and Jenq-Neng
  Hwang.
\newblock Samurai: {Adapting} segment anything model for zero-shot visual
  tracking with motion-aware memory.
\newblock \emph{arXiv preprint arXiv:2411.11922}, 2024.

\bibitem[Zeng et~al.(2022)Zeng, Dong, Zhang, Wang, Zhang, and
  Wei]{zeng_motr_2022}
Fangao Zeng, Bin Dong, Yuang Zhang, Tiancai Wang, Xiangyu Zhang, and Yichen
  Wei.
\newblock Motr: {End}-to-end multiple-object tracking with transformer.
\newblock In \emph{ECCV}, pages 659--675. Springer, 2022.

\bibitem[Zhang et~al.(2023)Zhang, Gao, Xiao, and Fan]{zhang_animaltrack_2023}
Libo Zhang, Junyuan Gao, Zhen Xiao, and Heng Fan.
\newblock {AnimalTrack}: {A} {Benchmark} for {Multi}-{Animal} {Tracking} in the
  {Wild}.
\newblock \emph{IJCV}, 131\penalty0 (2):\penalty0 496--513, 2023.

\bibitem[Zhang et~al.(2022)Zhang, Sun, Jiang, Yu, Weng, Yuan, Luo, Liu, and
  Wang]{zhang_bytetrack_2022}
Yifu Zhang, Peize Sun, Yi Jiang, Dongdong Yu, Fucheng Weng, Zehuan Yuan, Ping
  Luo, Wenyu Liu, and Xinggang Wang.
\newblock Bytetrack: {Multi}-object tracking by associating every detection
  box.
\newblock In \emph{ECCV}, pages 1--21. Springer, 2022.

\bibitem[Zheng et~al.(2022)Zheng, Zhang, Zhang, Qi, and
  Sun]{zheng_progressive_2022}
Anlin Zheng, Yuang Zhang, Xiangyu Zhang, Xiaojuan Qi, and Jian Sun.
\newblock Progressive end-to-end object detection in crowded scenes.
\newblock In \emph{CVPR}, pages 857--866, 2022.

\bibitem[Zheng et~al.(2024)Zheng, Lin, Zuo, Fu, and Pan]{zheng_nettrack_2024}
Guangze Zheng, Shijie Lin, Haobo Zuo, Changhong Fu, and Jia Pan.
\newblock Nettrack: {Tracking} highly dynamic objects with a net.
\newblock In \emph{CVPR}, pages 19145--19155, 2024.

\bibitem[Zhu et~al.(2021)Zhu, Su, Lu, Li, Wang, and Dai]{zhu_deformable_2021}
Xizhou Zhu, Weijie Su, Lewei Lu, Bin Li, Xiaogang Wang, and Jifeng Dai.
\newblock Deformable detr: Deformable transformers for end-to-end object
  detection.
\newblock In \emph{ICLR}, 2021.

\end{thebibliography}
}
\appendix
\clearpage
\setcounter{page}{1}
\maketitlesupplementary

\section{Dataset statistics}
\label{sec:appendix_dataset}
 The main characteristics of the datasets used in our experiments are summarized in \autoref{tab:dataset_statistics}. These datasets cover a wide variety of species, environments, and tracking challenges, from sparse scenes with few individuals to dense, dynamic groups. The statistics illustrate differences in scale, sequence length, and tracking complexity, providing essential context for interpreting our results.

\begin{table}[!h]
  \caption{\textbf{Detailed dataset statistics.} Overview of dataset properties and tracking challenges across evaluated benchmarks.}
  \label{tab:dataset_statistics}
  \centering
  \footnotesize
  \setlength{\tabcolsep}{3pt}
  \begin{tabular}{lcccc}
    \toprule
      & ChimpAct & BFT & AnimalTrack & GMOT-40\\
    \midrule
    Splits          & trn/val/tst & trn/val/tst & trn/tst   & tst       \\
    Sequences       &   163       &   106       &    58      &  16       \\
    Frames          &   160.8K    &  19.3K      &   24.7K    &   3.7K    \\
    Classes         &     1       &    1        &    10      &   4       \\
    FPS             &    25       &   25        &    30      &  24--30   \\
    Resolution      & 576--720p   & 1080p       & 1080--1440p& 1080p     \\
    \midrule
    Min.\ len.\ (s)&   22.9      &    0.9      &   6.5      &   3.3     \\
    Avg.\ len.\ (s)&   39.5      &    7.3      &  14.2      &   7.7     \\
    Max.\ len.\ (s)&   40.0      &   22.2      &  75.6      &  24.3     \\
    Total len.\ (s)& 6432.8      &  773.2      & 823.7      & 123.1     \\
    \midrule
    Avg.\ tracks    &     4       &    6        &   33       &   61      \\
    Max.\ tracks    &     9       &   40        &  134       &  128      \\
    Total tracks    &   712       &  668        & 1{,}927    &  980      \\
    Total boxes     &   563K      &   85K       &  429K      &   97K     \\
    \bottomrule
  \end{tabular}
\end{table}

\section{Extended metrics}
\label{sec:appendix_metrics}
To provide deeper insight into our model's performance and facilitate comparison with prior work, we report additional metrics on the test sets (\autoref{tab:extended_metrics}).

\begin{table}[!h]
  \caption{\textbf{Performance on diverse multi-animal tracking benchmarks.} We report standard MOT metrics to assess detection accuracy, association consistency, and overall tracking robustness.}
  \label{tab:extended_metrics}
  \centering
  \footnotesize
  \setlength{\tabcolsep}{2.5pt}
  \begin{tabular}{@{}lcccccccc@{}}
    \toprule
    Dataset & \rotatebox{60}{HOTA$\uparrow$} & \rotatebox{60}{DetA$\uparrow$} & \rotatebox{60}{AssA$\uparrow$} & \rotatebox{60}{DetRe$\uparrow$} & \rotatebox{60}{LocA$\uparrow$} & \rotatebox{60}{MOTA$\uparrow$} & \rotatebox{60}{IDF1$\uparrow$} & \rotatebox{60}{IDSW$\downarrow$}\\
    \midrule
    ChimpAct       & 58.6 & 49.8 & 70.1 & 57.3 & 83.4 & 48.6 & 66.7 &  32  \\
    BFT            & 74.8 & 72.2 & 77.7 & 80.5 & 87.8 & 81.8 & 88.4 &  51  \\
    AnimalTrack    & 58.0 & 52.7 & 65.2 & 63.8 & 81.1 & 58.9 & 72.0 & 442  \\
    GMOT-40        & 62.4 & 57.2 & 69.2 & 67.2 & 80.1 & 64.7 & 77.4 & 496  \\
    \bottomrule
  \end{tabular}
\end{table}

\section{Runtime and memory requirements}
\label{sec:appendix_runtime}
\begin{figure}[htb]
  \centering
  \includegraphics[width=\linewidth]{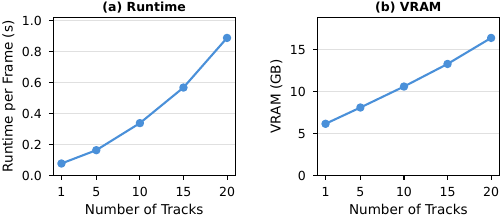}
  \caption{\textbf{Runtime and VRAM requirements for different numbers of active tracks.} The inference speed and memory consumption of SAM 2 increase with the number of tracked objects, indicating limited scalability in crowded scenes.}
  \label{fig:runtime}
\end{figure}

The runtime per image and VRAM requirements for SAM 2 Hiera-L are shown in \autoref{fig:runtime}. All measurements are performed on a single NVIDIA A100. Both runtime and VRAM consumption scale at least linearly with the number of tracked objects (\autoref{fig:runtime}), making the tracker less suitable for crowded scenarios. This limits the model's applicability on edge devices, making it more suitable for generating pseudo-labels or operating on small-scale data with no or few labels.

VRAM usage is measured using \texttt{torch.cuda.memory.memory\_allocated}. Runtime is measured over a 300-frame sequence and includes image loading, encoding, video propagation, and heuristic operations. Longer sequences increase VRAM because SAM 2 stores all image embeddings in GPU memory. Image embeddings and track states can be offloaded to the CPU to reduce VRAM at the expense of higher runtime.

\section{Failure cases}
\label{sec:failure_cases}
Missed detections remain our primary error source, as reflected by DetA falling substantially below AssA in \autoref{tab:extended_metrics}. Errors also arise during tracking itself. Because SAM~2 weighs appearance more heavily than motion, tracks occasionally jump to a visually similar individual when the target is occluded or leaves the scene. In crowded scenes, multiple individuals are occasionally merged into a single track. The model is also not designed for long-term re-identification. Tracks are terminated after 30 frames without a detection, and even with this threshold relaxed, SAM~2 lacks the individual-identity supervision required for reliable long-term association.

\begin{figure*}[tb]
    \centering
    \includegraphics[width=\linewidth]{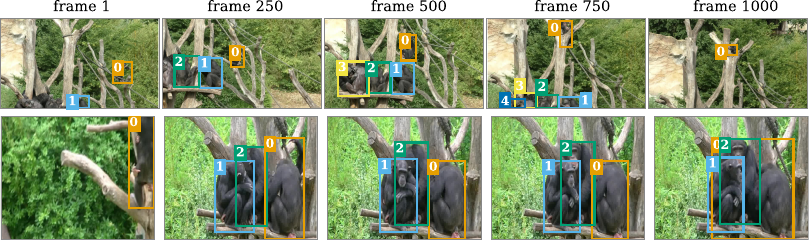}
    \caption{\textbf{Visualization of predictions on ChimpAct.} The predicted bounding boxes as well as the track id (distinguished by color and number) are visualized for two ChimpAct sequences from the test set.} 
    \label{fig:vis}
\end{figure*}

\section{Stress conditions covered by the benchmarks}
\label{sec:appendix_stress}
While we do not run dedicated stress tests, the four benchmarks span complementary challenging conditions. ChimpAct probes occlusion and long sequence durations (up to 40\,s with frequent mutual occlusion). BFT probes fast motion and appearance deformation across various bird species in flight. AnimalTrack and GMOT-40-Animal probe density and visual ambiguity, with up to 134 and 128 simultaneous tracks, respectively, across diverse animal categories. Consistent performance across all four indicates that our modifications generalize across these conditions without per-dataset tuning.

\section{Visualization}
\label{sec:vis}
The predicted bounding boxes are visualized for two sequences of the ChimpAct test set (\autoref{fig:vis}).

\section{Hyperparameters}
\label{sec:appendix_hyperparam}
This section provides an overview of the text prompts and hyperparameter settings used in all experiments. The text prompts correspond to the expected object categories present in each dataset and are used as input to Grounding DINO (\autoref{tab:text_prompts}). To ensure a fair comparison with SAM 3, we verified that the chosen prompts are compatible with its text encoder. In cases where a prompt led to degraded SAM 3 performance due to encoder sensitivity, we substituted it with a suitable alternative (e.g., \texttt{livestock} instead of \texttt{cow} for GMOT-40-Animal). The hyperparameters include all values related to object addition, reconstruction, and cross-object interaction, along with their corresponding explanations (\autoref{tab:hyperparameters}). All experiments are conducted using the same parameter configuration for every dataset.

\begin{table}[!h]
  \caption{\textbf{Text prompts used for each dataset.} The same set of prompts is provided to Grounding DINO across all experiments for the respective dataset.}
  \label{tab:text_prompts}
  \centering
  \footnotesize
  \begin{tabular}{@{}ll@{}}
    \toprule
    Dataset & Text prompts\\
    \midrule
    ChimpAct & ape\\
    BFT & bird\\
    AnimalTrack & chicken, deer, dolphin, duck, goose,\\ 
    &  horse, penguin, pig, rabbit, zebra\\
    GMOT-40 & bird, fish, insect, cow, sheep, goat, wolf\\
    \bottomrule
  \end{tabular}
\end{table}

\section{Detection score distribution}
\label{sec:appendix_dist}
\begin{figure*}
    \centering
    \includegraphics[width=\linewidth]{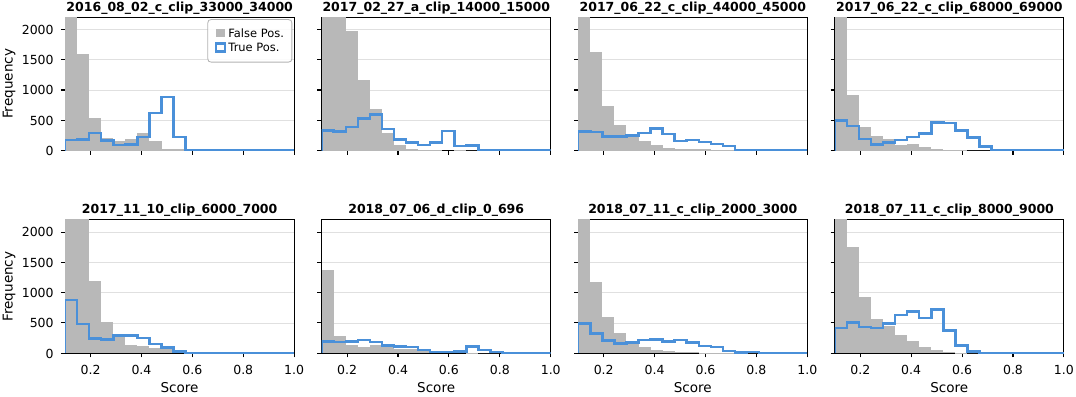}
    \caption{\textbf{The distribution differs significantly between sequences}. Detection score distribution from different sequences of the ChimpAct test split. }
    \label{fig:dist_seq}
\end{figure*}

Detection score distributions differ markedly between in-domain and zero-shot detectors and also vary substantially between individual sequences (\autoref{fig:dist_seq}).

\begin{table*}[!t]
  \caption{\textbf{Hyperparameter settings used in all experiments.} The same values were applied across all datasets. Hyperparameters in bold were newly introduced by us. The other hyperparameters and their values were taken from SAM2MOT~\cite{jiang_sam2mot_2026}.}
  \label{tab:hyperparameters}
  \centering
  \footnotesize
  \begin{tabular}{llc}
    \toprule
    Hyperparameters & Description & Value\\
    \midrule
    \multicolumn{3}{l}{\textit{Object addition}}\\
    $\boldsymbol{\delta}$ & Static offset added to adaptive detection threshold & 0.1\\
    $\boldsymbol{\tau_\text{mask}}$ & Maximum intersection between existing mask and new mask & 0.4\\
    \midrule
    \multicolumn{3}{l}{\textit{Object reconstruction \& removal}}\\
    $\boldsymbol{\tau_\text{IoU}}$ & Minimum difference between first and second detection-track IoU & 0.3\\
    $\tau_\text{reliable}$ & Tracks with an occlusion score above are assumed to be reliable & 8\\
    $\tau_\text{pending}$ & Tracks with an occlusion score above are assumed to be pending & 6\\
    $\tau_\text{lost}$ & Tracks with an occlusion score below are assumed to be lost & 2\\
    $N_\text{lost}$ & Number of consecutive frames after which a lost track is terminated & 25\\
    \midrule
    \multicolumn{3}{l}{\textit{Cross-object interaction}}\\
    $N_\text{frames}$ & Number of frames used to calculate the standard deviation & 10\\
    $\tau_\text{mIoU}$ & Minimum mask overlap for cross-object interaction & 0.8\\
    $\tau_{\Delta \text{score}}$ & Minimum occlusion score difference for cross-object interaction & 2\\
    $\tau_{\Delta\text{std}}$ & Minimum occlusion score std difference for cross-object interaction & 0.2\\
    \midrule
    \multicolumn{3}{l}{\textit{Non-maximum suppression}}\\
    $\boldsymbol{\tau_\text{nms}}$ & Mask IoU threshold for non-maximum suppression & 0.95\\
    \bottomrule
  \end{tabular}
\end{table*}

\end{document}